\documentclass[conference]{IEEEtran}
\IEEEoverridecommandlockouts
\usepackage{cite}
\usepackage{amsmath,amssymb,amsfonts}
\usepackage{algorithmic}
\usepackage{graphicx}
\usepackage{textcomp}
\usepackage{xcolor}
\def\BibTeX{{\rm B\kern-.05em{\sc i\kern-.025em b}\kern-.08em
    T\kern-.1667em\lower.7ex\hbox{E}\kern-.125emX}}
\begin{document}

\title{Real-time object detection and robotic manipulation for agriculture using a YOLO-based learning approach\\
}


\author{Hongyu Zhao, Zezhi Tang$^{\ast}$, Zhenhong Li, Yi Dong, Yuancheng Si, Mingyang Lu, George Panoutsos

\thanks{H. Zhao is with the Department of Physics, Imperial College, London, United Kingdom (email: hz2623@ic.ac.uk)}%
 \thanks{Z. Tang and G. Panoutsos are with the Department of Automatic Control and Systems Engineering, University of Sheffield, Sheffield, S1 3JD, United Kingdom (emails: zezhi.tang@sheffield.ac.uk,  g.panoutsos@sheffield.ac.uk)}\thanks{Z. Li is with the Department of Electrical and Electronic Engineering, University of Manchester, Manchester, United Kingdom (email: zhenhong.li@manchester.ac.uk)}
\thanks{Y. Dong is with the Department of Electronics and Computer Science, University of Southampton, Southampton, United Kingdom (email: yi.dong@soton.ac.uk)}
\thanks{Y. Si is with the Department of Economics, Fudan University, Shanghai,  200433, China (email: siyuancheng@fudan.edu.cn)}
\thanks{M. Lu is with the Center for Nondestructive Evaluation (CNDE), Iowa State University, Ames, IA 50011, United States (email: mingylu@iastate.edu)}
\thanks{*Corresponding author}
\thanks{*© 2024 IEEE.  Personal use of this material is permitted.  Permission from IEEE must be obtained for all other uses, in any current or future media, including reprinting/republishing this material for advertising or promotional purposes, creating new collective works, for resale or redistribution to servers or lists, or reuse of any copyrighted component of this work in other works.}

}

\maketitle

\begin{abstract}

The optimisation of crop harvesting processes for commonly cultivated crops is of great importance in the aim of agricultural industrialisation. Nowadays, the utilisation of machine vision has enabled the automated identification of crops, leading to the enhancement of harvesting efficiency, but challenges still exist. This study presents a new framework that combines two separate architectures of convolutional neural networks (CNNs) in order to simultaneously accomplish the tasks of crop detection and harvesting (robotic manipulation) inside a simulated environment. Crop images in the simulated environment are subjected to random rotations, cropping, brightness, and contrast adjustments to create augmented images for dataset generation. The \textit{you only look once} algorithmic framework is employed with traditional rectangular bounding boxes $($R-Bbox$)$ for crop localization. The proposed method subsequently utilises the acquired image data via a visual geometry group model in order to reveal the grasping positions for the robotic manipulation.

\end{abstract}


\section{INTRODUCTION}

The progression of automation can be observed on a global scale across several industries. The modernization and automation of agricultural production have also been noticed. The implementation of mechanized techniques in agriculture has facilitated the automation of diverse processes, resulting in enhanced efficiency in agricultural production \cite{a2}. Nevertheless, the issue pertaining to crop harvesting within the realm of automation remains inadequately resolved, with conventional robots encountering challenges in accurately perceiving and successfully executing the act of crop grasping.

Traditional machines have faced challenges when it comes to harvesting crops. Manual labor is time-consuming and leads to increased production costs. Therefore, robots can be used to contribute to increased agricultural productivity \cite{a0}. In the context of industrial production lines, robots typically perform specific roles within a production task, such as the manipulation and placement of products at a fixed location or the execution of specific steps within a specialized process\cite{b1}. Nevertheless, using robots in agricultural productivity requires enhanced object detection and grasping capabilities. Hence, it is necessary to conduct research on the topic of robot recognition and grasping techniques.
Specifically, the concept of grasping holds significant importance in the field of automation, as the majority of automated systems rely on the precise and effective gripping of a designated object. Currently, a wide range of algorithms exist that are used for the purpose of object recognition in conjunction with robotic grasping. 

The mask region-based convolutional neural network (Mask-RCNN) algorithm is employed for the purpose of segmenting and performing geometric stereo-matching in order to accurately determine the location of the object of interest in the camera's field of view; the robot manipulator subsequently performs efficient grabbing of the target object\cite{z1}. The grasp region-based convolutional neural network(GR-ConvNet) algorithm has the capability to produce grasping poses based on RGB photos. This addresses the challenge of creating and performing grasping actions for a robot that is unfamiliar with the items in its environment, using n-channel photographs of the scene\cite{z2}.

The YOLO method is a computer vision technique utilised for object recognition, renowned for its remarkable real-time detection capabilities. Moreover, it is continuously subjected to optimisation and enhancement efforts. As discussed in \cite{b2}, the utilisation of YOLO effectively addresses the significant challenge of domain drift commonly seen in traditional target detection approaches. This enables the generalised underwater object detector (GUOD) to achieve commendable performance in detecting targets in diverse underwater settings. The authors of \cite{b8} propose a solution to address the challenge of picture detection on datasets with limited samples. The proposed approach uses the real-time capabilities of YOLO, along with techniques such as transfer learning and data augmentation, to enhance detection rates and speed. In \cite{x1}, the YOLO is acknowledged for addressing the difficulty of accurately detecting irregular items in complicated situations. The proposed approach enhances the speed of detection and improves the likelihood of successfully capturing such objects. 

Meanwhile, while YOLO is known for its rapid and effective real-time detection skills, the VGG algorithm has distinct advantages in the fields of deep learning and feature extraction. The VGG models, particularly their deep architecture, can effectively capture intricate and complex image characteristics. This attribute proves to be particularly significant when dealing with visual sceneries of a highly complex nature. In \cite{b7}, the authors employed the VGG architecture to segment CT images precisely. It is imperative to develop a segmentation approach for mass spectrometry that is both accurate and efficient in order to support clinical applications. In \cite{x2}, the utilisation of the VGG model employed in this research transmits the target object's geometric centre coordinates and long-side angle details to the manipulator. Consequently, the manipulator can carry out the classification and grasping of the target object. This approach effectively facilitates the implementation of a garbage classification control method. Therefore, integrating YOLO's real-time detection capacity with VGG's proficiency in deep features enables the development of a visual recognition system with enhanced power and accuracy.

In this paper, the models will be trained in a simulated environment because data obtained from a simulated world remains unaffected by external interference, allowing for the convenient acquisition of high-quality datasets. Two improved CNN architectures, YOLO and VGG, will be employed. Compared to traditional algorithms that solely focus on image recognition, the advantage of the approach used in this paper lies in integrating two distinct neural architecture types to enable precise harvesting by robots. This approach goes beyond traditional techniques that only use the YOLO neural framework. It identifies the objects to be harvested and involves the VGG model in further processing the images of the objects determined by the YOLO model. This combined approach allows the model to calculate the optimal gripping points, enhancing the precision of harvesting.

\section{RELATED WORK}

The existing methodology is based on the utilization of CNNs due to their exceptional performance in large-scale image recognition tasks. Due to the proven efficacy of CNNs, developers have extensively adopted object identification algorithms. Investigating the complexity of CNN construction helps in understanding the advantages it offers in the realm of image recognition.

\subsection{The Principles of CNN}

The CNN model is inspired by biological vision, particularly the feedforward neural network architecture with a convolutional structure. The key differentiating factor between this approach and conventional methods is its ability to learn and extract features from the given dataset automatically\cite{a7}. CNN possesses the advantage of local connectivity, as depicted in Fig. 1. In contrast to connecting with every neuron in the preceding layer, each neuron in a CNN is only linked to neurons within a local region of the input\cite{a8}. This reduces the number of parameters in the network, as each neuron is responsible for processing only a portion of the data. Additionally, it accelerates the information flow within the network as the connections between neurons are more sparse\cite{a9}. The main difference between fully connected layers and CNNs is that each node in the former is connected to all nodes in the previous layer, and each connection has a corresponding weight\cite{a7}. This architecture enables fully connected layers to learn complex patterns and relationships in the input data. However, it leads to an increase in computational complexity.

\begin{figure}[htbp]
\centering
\includegraphics[width=0.4\textwidth]{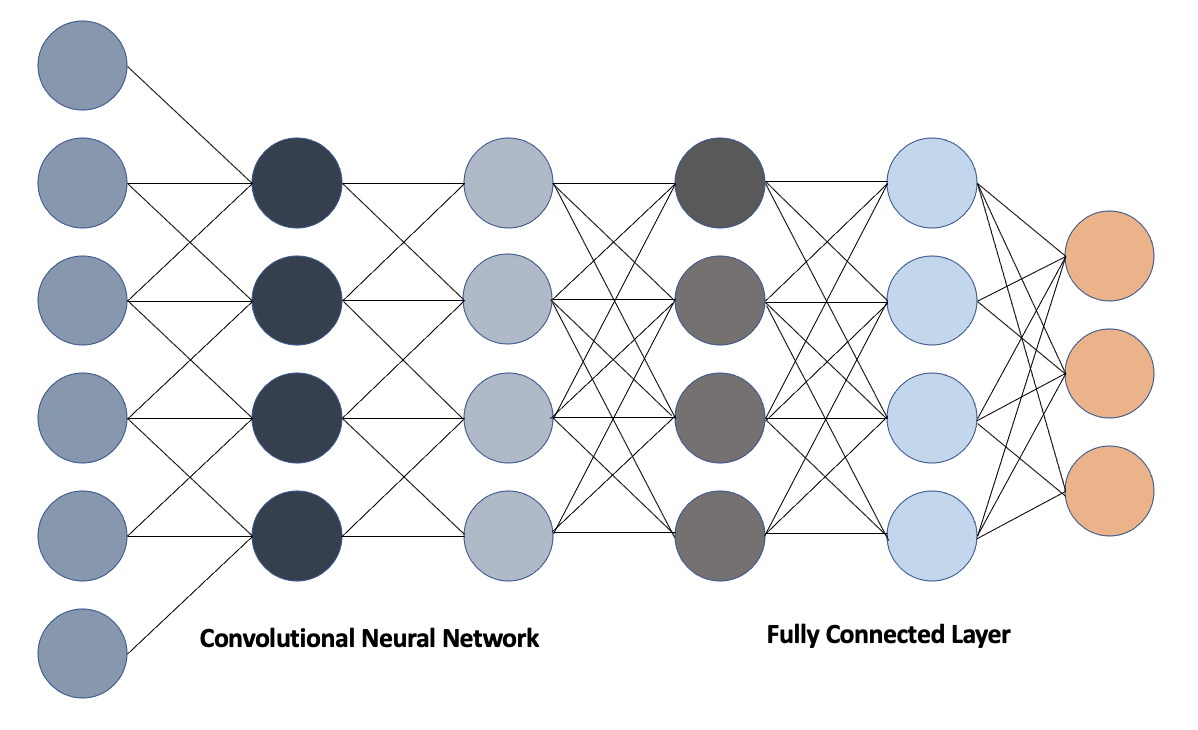}
\caption{CNN and FC structure.}
\label{fig}
\end{figure}

\begin{figure}[htbp]
\centering
\includegraphics[width=0.5\textwidth]{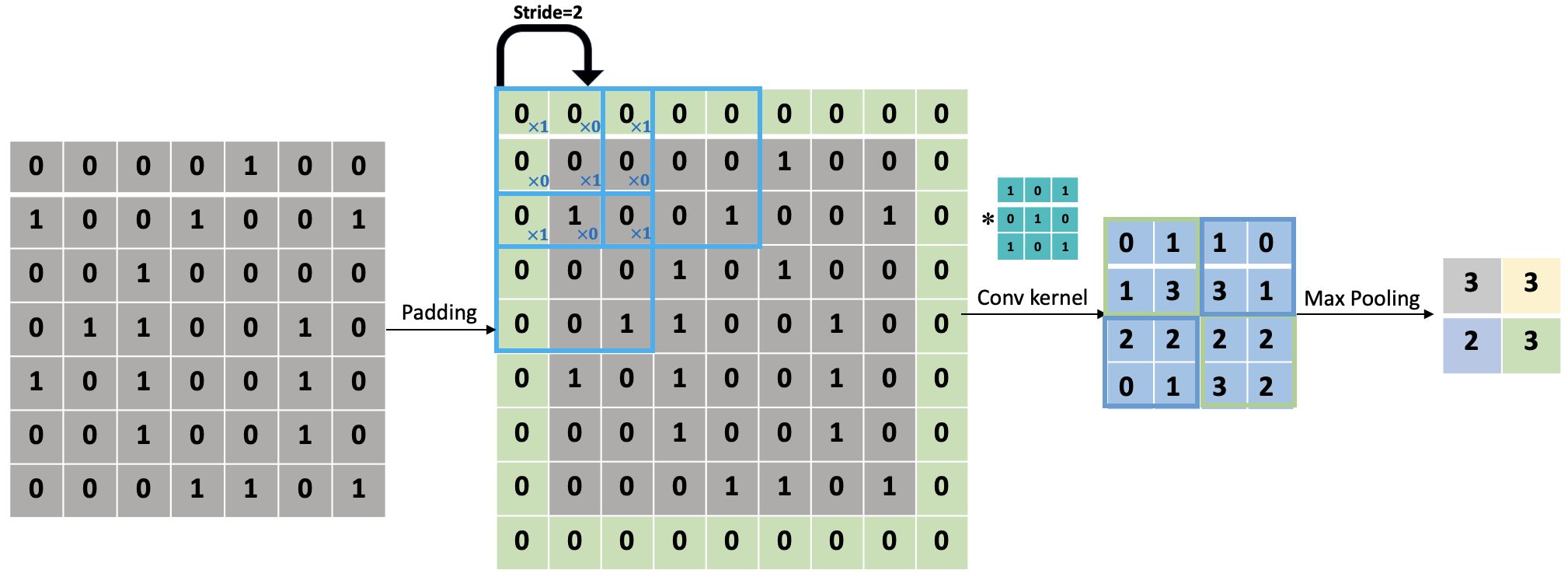}
\caption{Procedure of 2-D CNN.}
\label{fig}
\end{figure}

CNNs typically include a pooling layer, often positioned after the convolutional layer as depicted in Fig. 2. This layer reduces the size of the feature map, effectively reducing its width and height and thereby decreasing the volume of data and computational complexity\cite{a10}. Through the pooling operation, which aggregates information from neighboring pixels, the essential features of the image are retained while reducing the dimensionality of data. Ultimately, this process trims the number of parameters by eliminating unimportant image features\cite{a12}.

\subsection{YOLO Architecture and Loss Function}

\subsubsection{YOLO Architecture}


The main difference between YOLO and conventional systems is the simultaneous processing of bounding boxes and object classification within YOLO. The input image is initially partitioned into a grid of $S*S$ cells, with each cell bearing A bounding boxes.  A confidence score is assigned to each bounding box, indicating the likelihood of an object being present within that specific bounding box, which is denoted by

\begin{equation}\label{CriticNN}
    \begin{aligned}
        C = Pr(Object)*IOU_{pred}^{truth},
    \end{aligned}
\end{equation}
where confidence score $C$ is calculated as $C = Pr(Object)*IOU_{pred}^{truth}$ ,$0\le C \le 1 $, which is used to assess the quality of bounding box predictions. $Pr(Object)$ represents the probability of an object's presence in a bounding box, and $IOU_{pred}^{truth}$ measures the overlap between the predicted and true bounding boxes, indicating the accuracy of the prediction, and $0 \le Pr(Object) IOU_{pred}^{truth} \le 1 $.

Simultaneously, while generating bounding boxes, each grid cell assesses the probability of the presence of a particular class of object. The specific probability within each grid cell is defined as:
\begin{equation}\label{CriticNN}
    \begin{aligned}
        &Pr(Class_{i}|Object)*Pr(Object)*IOU_{pred}^{truth}\\
    &= Pr(class_{i})*IOU_{pred}^{truth},  
    \end{aligned}
\end{equation}
where $Pr(Class_{i}|Object)$ is the conditional probability that an object belongs to a specific class, $Class_{i}$, given that an object is detected. It represents the likelihood that the object in a specific grid cell belongs to class $i$.
However, many of these predicted bounding boxes have low confidence scores. A certain threshold is set to filter out most of these low-confidence bounding boxes. In target detection tasks, multiple overlapping bounding boxes are generated\cite{a13}. To select the most suitable bounding boxes and reduce redundant detection results while improving accuracy, the Non-Maximum Suppression (NMS) technique is employed. NMS is used to discard all bounding boxes in a class with confidence levels lower than a given threshold. For the remaining bounding boxes, pairwise comparisons are performed based on IoU\cite{a14}. If the IoU between two bounding boxes exceeds 0.5, the one with lower confidence is removed. Otherwise, both are retained in the list. This process effectively sparsifies similar bounding boxes, resulting in the selection of the most appropriate bounding box.

\subsubsection{YOLO Loss Function}

The YOLO method employs the following formula to calculate loss and optimize confidence:

\begin{equation}\label{CriticNN}
    \begin{aligned}
        \text{Loss} = 
        &\quad \lambda_{\text{coord}} \sum_{i=0}^{s^{2}} \sum_{j=0}^{B} \delta_{ij}^{\text{obj}} \left [ \left ( x_{i} -\hat{x}_{i} \right )^{2}+ \left ( y_{i}-\hat{y_{i}} \right )^{2} \right ] \\
        &\quad + \lambda_{\text{coord}} \sum_{i=0}^{s^{2}}\sum_{j=0}^{B} \delta_{ij}^{\text{obj}} \left [ \left ( \sqrt{w_{i}} -\sqrt{\hat{w}_{i}} \right )^{2}+ \left (h_{i}-\hat{h_{i}} \right )^{2} \right ] \\
        &\quad + \sum_{i=0}^{s^{2}}\sum_{j=0}^{B} \delta_{ij}^{\text{obj}} \left (C_{i} -\hat{C}_{i} \right )^{2} \\
        &\quad + \lambda_{\text{coord}} \sum_{i=0}^{s^{2}}\sum_{j=0}^{B} \delta_{ij}^{\text{obj}} \left (C_{i} -\hat{C}_{i} \right )^{2} \\
        &\quad + \sum_{i=0}^{s^{2}}\delta_{i}^{\text{obj}} \sum_{c \in \text{classes}} \left (p_{i}(c) -\hat{p}_{c}(c) \right )^{2},
    \end{aligned}
\end{equation}
where $Loss$ is the value that the YOLO model tries to minimize during training. It consists of several parts, each of which relates to a different aspect of detection. $\lambda_{coord}$ is a weight parameter used to increase the emphasis on the accuracy of the bounding box coordinates prediction, $B$ indicates the number of bounding boxes predicted per grid cell, $\delta_{ij}^{obj}$ is an indicator function, which is $1$ if an object is present in the $j$th bounding box of the  $i$th grid cell, and $0$ otherwise. $x_{i}$,$y_{i}$,$w_{i}$,$h_{i}$
are the actual bounding box coordinates, $\hat{x}_{i}$,$\hat{y}_{i}$,$\hat{w}_{i}$,$\hat{h}_{i}$ are the bounding box coordinates predicted by the model. $c_{i}$, $\hat{c}_{i}$ is the actual class probability and class probability predicted by the model.
$pi(c)$,$\hat{p}_{c}(c)$ is the actual class probability and class probability predicted by the model.

\subsection{VGG Loss Function}

The VGG model will reorganize the image into $224*224*3$ RGB format. After that, VGG will then be propagated forward through the neural network while calculating the loss value by applying the following formula:

\begin{equation}\label{CriticNN}
    \begin{split}
        Loss_{i} = &{ \sum_{i=1}^{N}}  \Big[ (x_{i1}- \hat{x} _{i1} )^{2} +(x_{i2}- \hat{x} _{i2} )^{2} \\
        &+(y_{i1}- \hat{y} _{i1} )^{2} +(y_{i2}- \hat{y} _{i2} )^{2} \Big] +\lambda{ \sum_{j=1}^{M}}w_{j},
    \end{split}
\end{equation}

where $Loss_{i}$ is the loss value for a single training sample, $N$ is the dataset of training samples, $M$ is weight parameters, $x_{i1}$, $x_{i2}$, $y_{i1}$, $y_{i2}$ are specific values from the actual image data, pixel values, color intensity. $\hat{x}_{i1}$, $\hat{x}_{i2}$, $\hat{y}_{i1}$, $\hat{x}_{i2}$ is the model's predicted values corresponding to the actual image data, $\lambda $  is a regularization parameter, $w$ is regularization term.

The equation is utilised in the VGG model to demonstrate the central aim, which is to minimise the average error between the predicted coordinates and the annotated values generated by the neural network. Additionally, a regularisation term is incorporated to address the issue of overfitting, which can impede the model's capacity to acquire novel information. During the training VGG model process, the original dataset is iteratively utilised, and the parameters of the neural network are modified according to the loss values. The optimisation of the backpropagation process is achieved by employing the Adaptive Moment Estimation (Adam) algorithm, which is a widely used optimisation technique for training neural networks. The learning rate is assigned a value of 0.1 in order to facilitate less frequent changes to the parameters.

\subsection{Simulation Environment}

The experiment employed the conventional UR5 model equipped with a vacuum gripper as the chosen gripper type. The decision to choose for alternative grippers was motivated by the project's emphasis on agricultural production, as conventional grippers have a higher propensity to cause harm to the skin of fruits and vegetables. The utilisation of a vacuum gripper effectively addresses this concern and improves the efficacy of gripping operations by simplifying the analysis to a solitary gripping point, hence minimising the computational complexity involved in grip calculations.

\begin{figure}[htbp]
\centering
\includegraphics[width=0.4\textwidth]{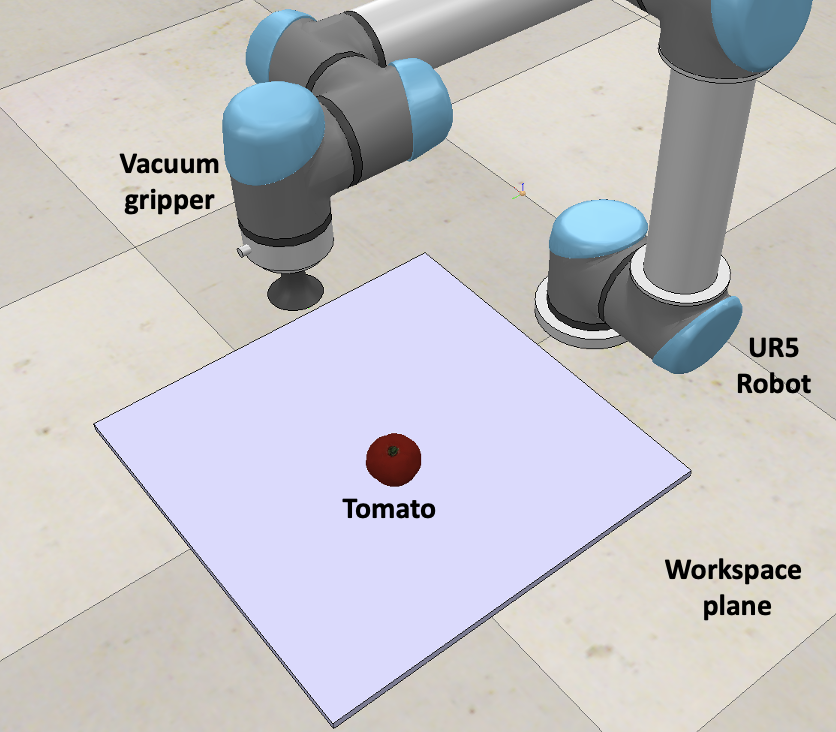}
\caption{Similation Environment}
\label{fig}
\end{figure}

\subsection{Data Labeling and Collection}
The objective of this experiment is to facilitate the robot's ability to categorise objects and autonomously ascertain appropriate gripping positions. The outcomes of model training are contingent upon the calibre of the dataset gathering. Hence, the data employed in this research is obtained from a simulated setting, wherein the integrity of the data is unaltered by outside factors such as noise and lighting, thereby guaranteeing the acquisition of data of superior quality.

The simulated environment is used to gather images of different postures of all crops which are shown in Fig. 3. These images are accompanied by labelled gripping point positions, and the related coordinate values are produced using the OpenCV library function. In order to enhance the trustworthiness of the overall experimental validation results and mitigate the risk of overfitting, we partitioned the entire dataset. The database is partitioned into training, validation, and testing sets using conventional machine-learning techniques. The training set is utilised to facilitate network training, while the validation set is employed to optimise the model by altering parameters. Finally, the testing set is used to validate the experimental outcomes and determine the final training results.

\section{Methodology}

\subsection{YOLO Structure}

The YOLO model is comprised of a structural arrangement which is shown in Fig. 4. that includes an initial layer responsible for transferring the input image to the backbone layer. This process involves the transformation of the input image into feature maps\cite {a13}.

\begin{figure}[htbp]
\centering
\includegraphics[width=0.5\textwidth]{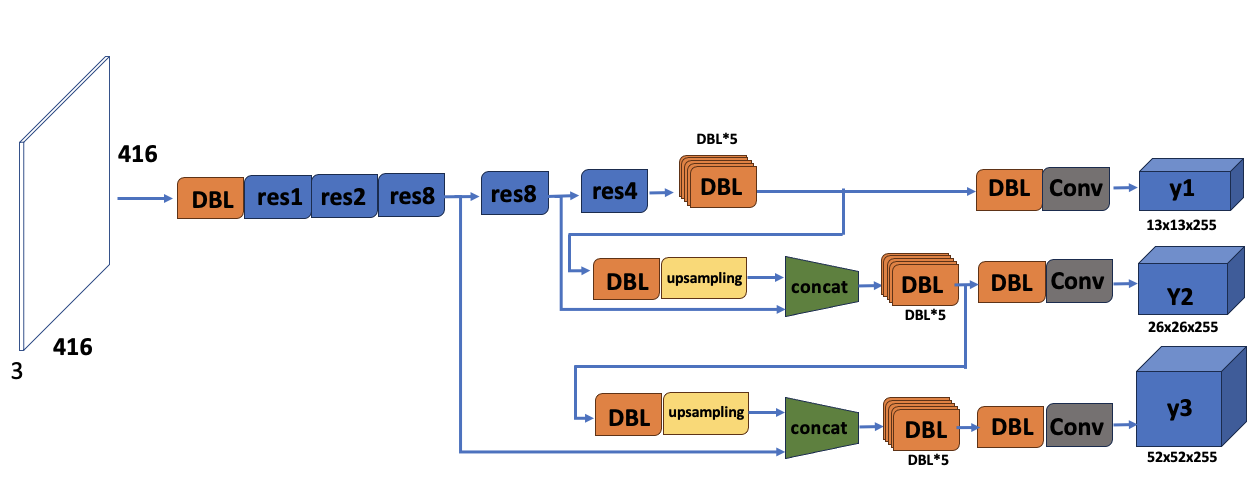}
\caption{YOLO Structure}
\label{fig}
\end{figure}

The development of YOLO aimed to create a process that incorporates both detection and classification. In contrast to conventional systems, YOLO enables simultaneous prediction of bounding boxes and object classification. The YOLO process involves a single evaluation of the input image, allowing for direct simultaneous prediction of bounding boxes and categories. The fastest YOLO framework can achieve up to 45 frames per second\cite{a13}. YOLO has trained the dataset that has been preprocessed through the construction of the network, initializing the weights and iterative training updates, optimizing the model through the loss function, and finally presenting a comparison of the data after the model has been applied.

\begin{figure}[htbp]
\centering
\includegraphics[width=0.45\textwidth]{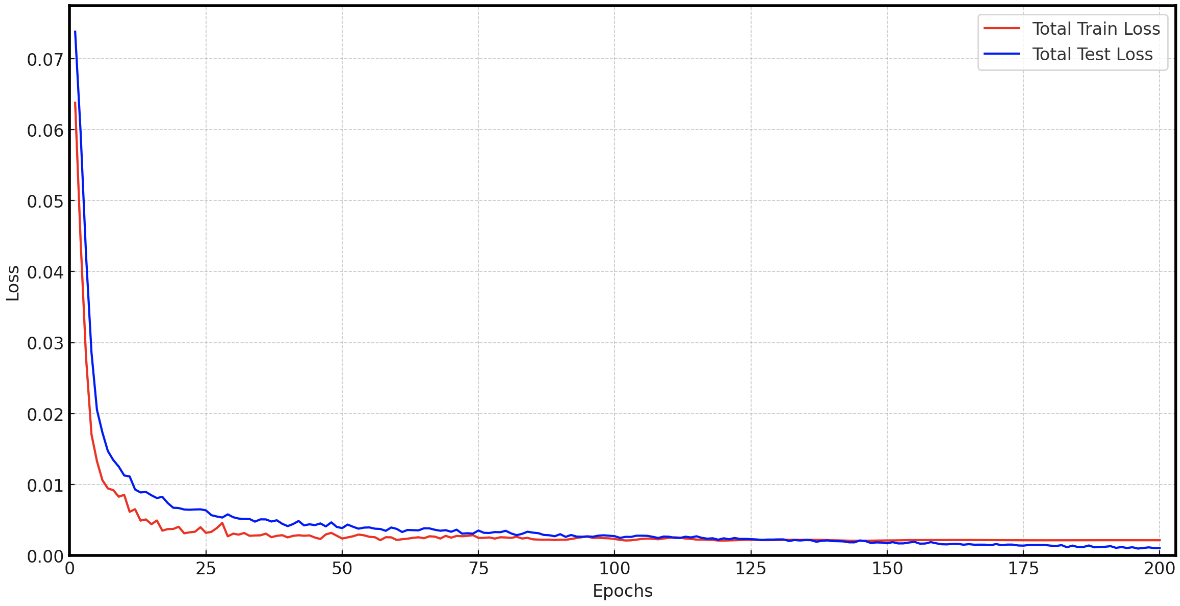}
\caption{Training Result of YOLO}
\label{fig}
\end{figure}

The analysis of the data reveals as depicted in Fig. 5. that as the duration of training increases, there is a corresponding decrease in the loss observed between the samples and the neural network. The rate of convergence exhibits considerable variability, ranging from rapid convergence to the attainment of a stable convergence state. The conclusive outcomes demonstrate a seamless training procedure, wherein the model's loss exhibits a consistent decline.

\subsection{VGG Structure}

This paper adopts the VGG  architecture, specifically VGG16, which consists of 13 convolutional layers and three fully connected layers as depicted in Fig. 6. 

\begin{figure}[htbp]
\centering
\includegraphics[width=0.45\textwidth]{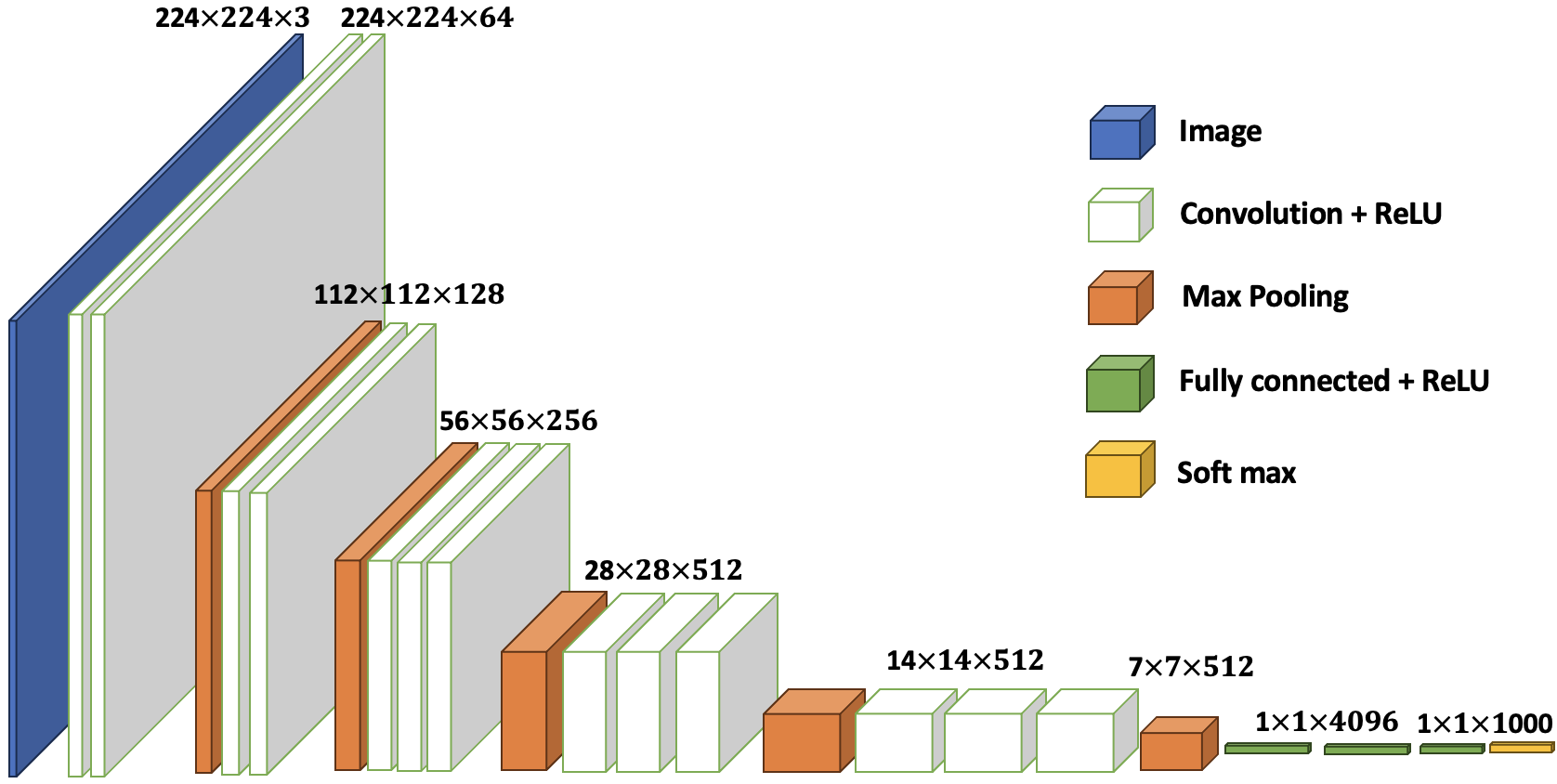}
\caption{Sturcture of Visual Geometry Group Network}
\label{fig}
\end{figure}

The initial phase of training the VGG model comprises a preprocessing step, wherein the mean value is subtracted from every image pixel. The architectural architecture of VGG16 is characterised by a series of convolutional layers that are then followed by pooling operations\cite{a16}. This pattern is iteratively replicated multiple times throughout the network. The architectural design presented herein enables the establishing of a hierarchical feature extraction network that is particularly suitable for recognition and classification tasks, it progressively absorbs increasingly complex and abstract attributes by integrating several convolution layers\cite{a17}.

The VGG architecture includes a nonlinear activation function within every convolutional layer, specifically the Rectified Linear Unit (ReLU). This inclusion facilitates the network's gathering and comprehension of complex nonlinear associations. By incorporating many convolutional layers, the neural network can gradually gain the capacity to construct complex feature representations. This enhancement in the network's expressive capability enables it to accommodate more intricate datasets effectively\cite{a18}.

\subsection{VGG Model}

In order to accurately identify the target position, it is necessary to employ a four-dimensional vector.   The matrix is acquired using the YOLO methodology, which generates the bounding rectangle.   Hence, it is essential to develop an algorithm that can effectively address this issue, as the accurate determination of a specific point requires the evaluation of a pair of coordinates, which requires considering four distinct values. In essence, the task at hand might be framed as a regression issue, with the objective of predicting these four specific values.   In order to tackle this difficulty, we utilise a VGG model for its resolution.

\begin{figure}[htbp]
\centering
\includegraphics[width=0.45\textwidth]{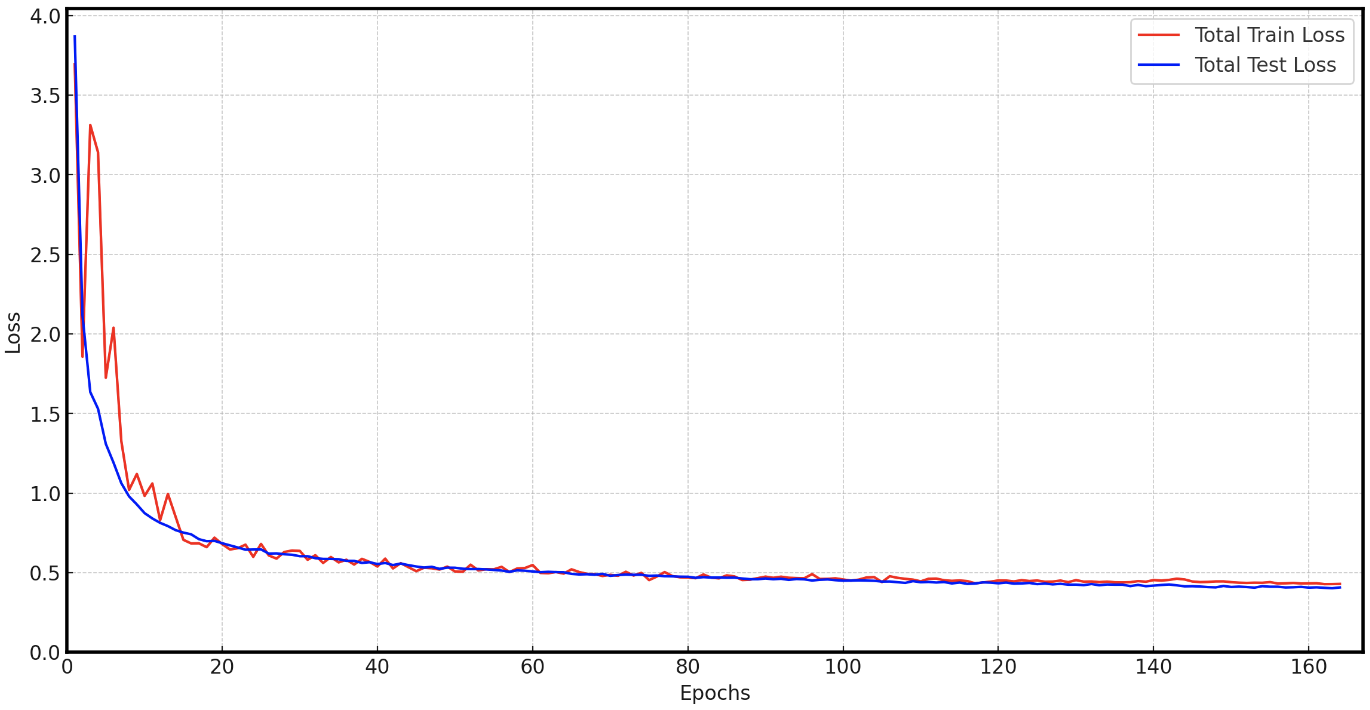}
\caption{Training result of VGG network}
\label{fig}
\end{figure}

The VGG model employed the preprocessed dataset and applied the same methodology for network development. The model was then iteratively trained to update its parameters, followed by optimisation through the application of a loss function. Ultimately, the construction of the model was successfully completed.

Upon comparing the data regarding the training process depicted in Fig. 7. it becomes evident that the loss values for both the training and test sets exhibit a decreasing trend. However, the variability observed in the initial phase of the test set is more wide. Nevertheless, as the training duration increases, there is a general trend towards stability. Both models exhibit convergence in their states and do not display signs of overfitting. Therefore, it can be concluded that the training results match the expected results.

\subsection{YOLO-VGG Architecture}

An efficient and accurate visual recognition system is successfully realized by combining the unique advantages of the YOLO and VGG models.  Based on its excellent real-time processing capability and efficient target detection performance, the YOLO model enables the system to recognize objects in an image quickly.  At the same time, the deep network architecture of the VGG model provides powerful feature extraction capabilities, enabling the system to capture finer and more complex image details.  This integrated approach improves the accuracy of recognition and enhances the system's ability to adapt to complex scenes.

\section{Result}
This section presents the data analysis of the training process for both models. Additionally, it demonstrates and analyses the operation of the two network designs in the simulation environment.

\subsection{Process Overview}

The combination of YOLO and VGG, along with the respective loss functions and network structures, is utilized using the preprocessed dataset for model construction. The objective is to enable the robot to achieve object classification and intelligent determination of the gripping point's position. Initially, the information captured by the camera in CoppeliaSim is reconstructed into a $416*416$ format. The YOLO neural network is then employed to process and classify the information and positions of each object in the image. The image is subsequently resized to $224*224$, and the collected image is fed into the VGG16 network to obtain the gripping position. This integration of the two techniques ensures that the combined approach aligns with the overall expectations of the experiment.

\subsection{YOLO-VGG Model Application}

After completing the training process, the YOLO neural network attempts to identify and classify placed objects in a simulated environment by examining the produced results, which are shown in Fig. 8.

\begin{figure}[htbp]
\centering
\includegraphics[width=0.3\textwidth]{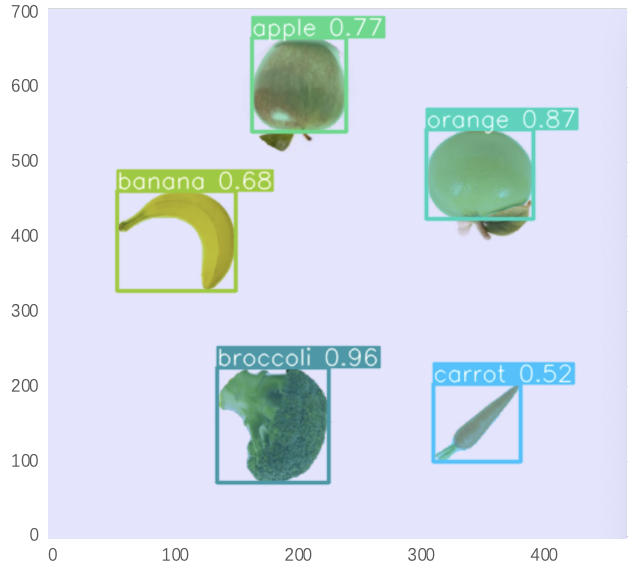}
\caption{Validity of YOLO test results}
\label{fig}
\end{figure}

The trained YOLO model demonstrates accurate object detection and classification in the given image by showing a high level of confidence. Additionally, the bounding box effectively contains the detected object, showing that the training method and outcomes match the experimental requirements.
Then we combine the proposed aforementioned YOLO scheme with a VGG model. Initially, the YOLO model performs object localization and category analysis. Subsequently, the image experiences a process of reconstruction and is resized to dimensions of $224$*$224$ pixels. The VGG model utilises the reconstructed image to choose an appropriate position for object gripping. Ultimately, the robot successfully acquires the thing by utilising the designated grabbing point.

\begin{figure}[htbp]
\centering
\includegraphics[width=0.5\textwidth]{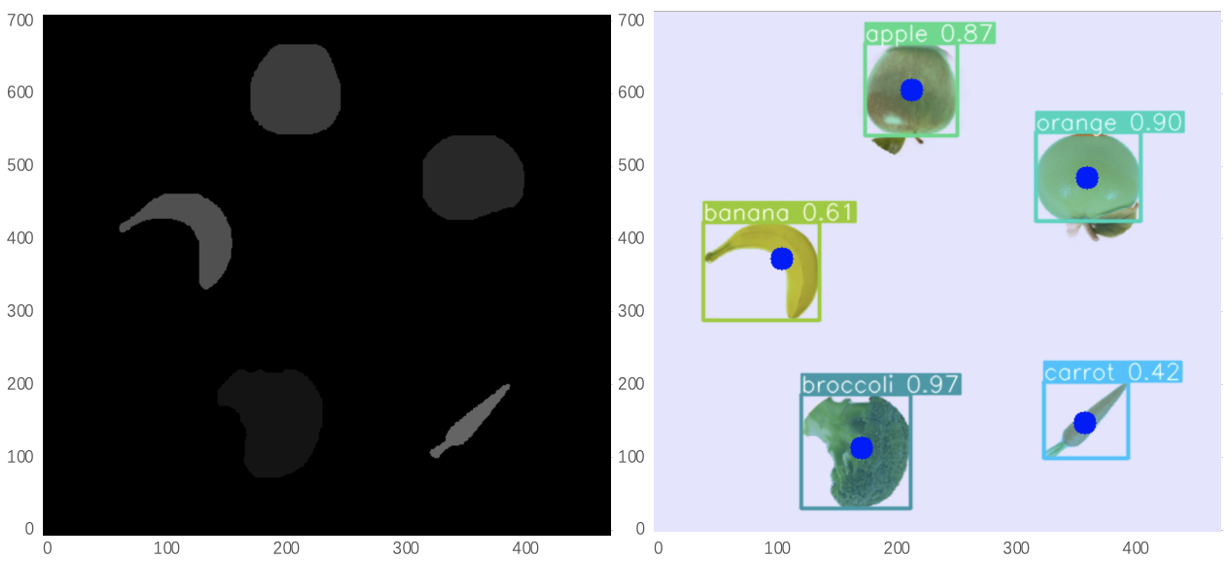}
\caption{Results of object position detection}
\label{fig}
\end{figure}

The provided visual representation shows the image data that has been processed by the VGG model, illustrating the outcome of two separate computational operations conducted by the VGG model. Firstly, a binarization method in machine vision reduces the features of an image, hence lowering computational complexity and assisting in the extraction of image features. Following this, the trained model is employed to calculate the existing optimal position of the centre of gravity for the placed object. The robot is able to effectively grasp the object by utilizing the positional data generated by the VGG network is depicted in Fig. 9. Even in the presence of ongoing changes in the object's location.

\section{Conclusion}

This paper presents a novel approach to object detection, combining the YOLO algorithm and the VGG16 architecture implemented in PyTorch to enhance robots' abilities in complex tasks like crop harvesting and classification.  Two neural networks were trained on a dataset for object recognition and precise prediction of optimal robot grasping positions.  The research demonstrates the regression network's high proficiency in accurately identifying and categorizing various crops, highlighting its robustness in object identification.  This suggests that further technological advancements could alleviate monotonous human labour in agricultural production.  Future research will aim to improve the robustness of the model to navigate foliage-rich environments and adapt to varying lighting conditions.


\end{document}